# The Z-axis, X-axis, Weight and Disambiguation Methods for Constructing Local Reference Frame in 3D Registration: An Evaluation

Bao Zhao, Xianyong Fang, Jiahui Yue, Xiaobo Chen, Xinyi Le, Chanjuan Zhao

*Abstract*—The local reference frame (LRF), as an independent coordinate system generated on a local 3D surface, is widely used in 3D local feature descriptor construction and 3D transformation estimation which are two key steps in the local method-based surface matching. There are numerous LRF methods have been proposed in literatures. In these methods, the x- and z-axis are commonly generated by different methods, and some x-axis methods are implemented on the basis of a z-axis being given. In addition, the weight and disambiguation methods are commonly used in these LRF methods. In existing evaluations of LRF, each LRF method is evaluated with a complete form. However, the merits and demerits of the z-axis, x-axis, weight and disambiguation methods in LRF construction are unclear. In this paper, we comprehensively analyze the z-axis, x-axis, weight and disambiguation methods in existing LRFs, and obtain six z-axis and eight x-axis, five weight and two disambiguation methods. Some axis methods are firstly proposed in this paper. The performance of these methods are comprehensively evaluated on six standard datasets with different application scenarios and nuisances. Considering the evaluation outcomes, the merits and demerits of different weight, disambiguation, z- and x-axis methods are analyzed and summarized. The experimental result also shows that some new designed LRF axes present superior performance compared with the state-of-the-art ones.

## I. Introduction

THREE dimensional surface matching is a foundational task in 3D computer vision and graphics with numerous applications in 3D object recognition [1, 2], 3D registration [3, 4], localization [5], etc. The surface matching can be solved by local and global methods [6]. Global methods [7] turn the geometric attributes of the whole model into a feature vector. However, it is sensitive to clutter and occlusion. Local methods [8-12] use local feature descriptors constructing point-to-point correspondences, which make these local methods have stronger robustness to clutter and occlusion than the global ones [6].

In the local method-based surface matching, local feature construction [10, 13, 14] and transformation estimation [15, 16] are the two key steps [17-19]. In the local feature construction, LRF is the basis for achieving the transformation invariance [20, 21], and in the transformation estimation methods, LRF is also an important tool for estimating a plausible transformation [15, 22]. Therefore, the performance of LRF is very important for designing superior local descriptor and transformation estimation methods. An LRF, as illustrated in Fig. 1, is an independent coordinate system established on a local surface, consisting of three orthogonal axes [23]. Its z-axis is named as local reference frame (LRA) [24]. The performance of LRF is mainly evaluated by two aspects: repeatability and robustness. An LRF is recognized as repeatable if its orientation variation is consistent with the transformation of the local surface [25]. The robustness indicates an ability to resist the impact of various nuisances including noise, occlusion, etc.

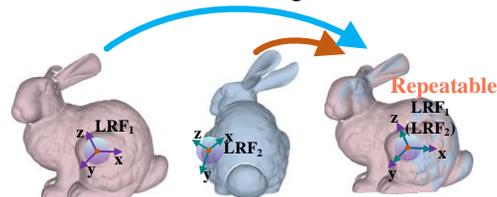

Fig. 1. An illustration of the repeatability for a LRF.

To improve the repeatability and robustness of an LRF, numerous LRF methods are proposed [23]. In these methods, most use the covariance analysis (CA) for constructing LRF [23]. These CA-based methods have two main steps [26]: The covariance matrix is firstly calculated to define the direction of the axes, and then the sign ambiguity of the axes is eliminated. Considering that the x-axis is more susceptible to occlusion than the z-axis [26], several geometrical attribute (GA)-based methods [23, 27] of separately constructing x-axis are proposed for strengthening the robustness of x-axis. After constructing the z- and x-axes, the y-axis is commonly generated by the cross of them. The details of the methods for constructing the z- and x-axes can be found in Sect. II.

For comparing the performance of existing LRFs, evaluations or reviews [23, 28] have been proposed. However, existing evaluations test the performance of each complete LRF method [23]. Since the x- and z-axis are commonly generated by different methods or strategies, and some x-axis methods are implemented on the basis of the z-axes [6, 26], these evaluations cannot accurately exhibit the merits and demerits of the z- and the x-axes in LRF. In addition, the weight and disambiguation methods are commonly used in existing LRF methods [23]. The performance of the weight and

---
Corresponding author: Bao Zhao.

Bao Zhao, Jiahui Yue and Chanjuan Zhao are with the School of Internet, Anhui university, Hefei, 230031, China (e-mail: zhaobao625@163.com, y259130@163.com, jojo20061864@126.com). Xianyong Fang is with the School of Computer Science and Technology, Anhui university, Hefei, 230031, China (e-mail: fangxianyong@ahu.edu.cn). Xiaobo Chen is with the School of Mechanical Engineering, Shanghai Jiao Tong University, Shanghai, 200240, China (e-mail: xiaoboc@sjtu.edu.cn). Xinyi Le is with the School of Electronic Information and Electrical Engineering, Shanghai Jiao Tong University, Shanghai 200240, China (e-mail: lexinyi@sjtu.edu.cn).

disambiguation methods are not comprehensively discussed in existing evaluations.

Motivated by these considerations, we analyze and summarize the methods of constructing z- and x-axes separately in existing LRF methods. Ultimately, six z-axis and eight x-axis methods are constructed. In these methods, some are firstly proposed in this paper. We also analyze the weight and disambiguation methods in existing LRF methods, and finally acquire five weight methods and two disambiguation strategies. For giving an unbiased evaluation of all axis methods, the support radius, weight and disambiguation of these axis methods, and the optimal z-axes in constructing x-axes are firstly evaluated and set. Then, the repeatability of all fourteen axis methods (including six z-axis and eight x-axis methods) are comprehensively evaluated on six popular datasets with different application scenarios and nuisances, and the robustness of these axis methods to eight common nuisances (e.g., Gaussian noise, varying mesh resolutions and occlusion) are evaluated. In addition, the influence of different axis methods to the performance of local feature descriptors and transformation estimation methods are also tested. Finally, the merits and demerits of these axis methods are summarized and discussed.

The main contributions of this paper are summarized as follows:

*First*, we analyze and summarize the methods of constructing z- and x-axes separately in existing LRFs, and comprehensively evaluate the repeatability and robustness of these z- and x-axis methods on six datasets with different application scenarios and nuisances.

*Second*, we summarize the weight and disambiguation methods in LRF construction and comprehensively evaluate their advantages and disadvantages.

*Third*, the merits and demerits of different weight, disambiguation, z- and x-axis methods are analyzed and summarized, which can provide a reference for researchers to design superior LRF methods.

*Fourth*, some superior axis methods (e.g., CA-M-b(z) and CA-M-b(x)) are firstly proposed in this paper.

The rest part of this paper is organized as follows. Section II presents the core ideas and computational steps of different weight, disambiguation, z- and x-axis methods. Section III gives an evaluation methodology. The experimental results are drawn in Section IV. Section V summarizes the merits and demerits of axis methods, and the conclusion is reported in Section VI.

## II. CONSIDERED METHODS

An complete LRF at a keypoint **p** is defined as a 3 × 3 matrix containing three orthogonal unit axes, i.e., $L(\mathbf{p})= [L(\mathbf{p}).\mathbf{x}\ L(\mathbf{p}).\mathbf{y}\ L(\mathbf{p}).\mathbf{z}]$, where $L(\mathbf{p}).\mathbf{x}$, $L(\mathbf{p}).\mathbf{y}$ and $L(\mathbf{p}).\mathbf{z}$ are 3 × 1 vectors storing unit vector coordinates. Commonly, the essence of the LRF construction is to calculate the $L(\mathbf{p}).\mathbf{x}$ and $L(\mathbf{p}).\mathbf{z}$, and then the $L(\mathbf{p}).\mathbf{y}$ is obtained via $L(\mathbf{p}).\mathbf{z} \times L(\mathbf{p}).\mathbf{x}$. Existing LRF methods are broadly categorized as: *covariance analysis* (CA)-based and *geometric attribute* (GA)-based methods. The general stages of the CA- and GA-based methods are presented in the Fig. 2. The CA-based method commonly includes three steps: 1) The covariance matrix of local points or mesh is calculated; 2) the eigenvector of the covariance matrix is computed as the direction of LRF axis; 3) the disambiguation of the axis direction is implemented to obtain the final LRF axis. The GA-based method employs local geometrical attributes (e.g., salient point) to calculate the axes of LRF, which is only used in the x-axis construction. In this paper, six CA-based methods for constructing z-axes and eight methods including five CA-based and three GA-based methods for generating the x-axes are considered. It is worth noting that some of these methods are firstly proposed in this paper. For readability, some common notations for describing these methods are listed in Table I.

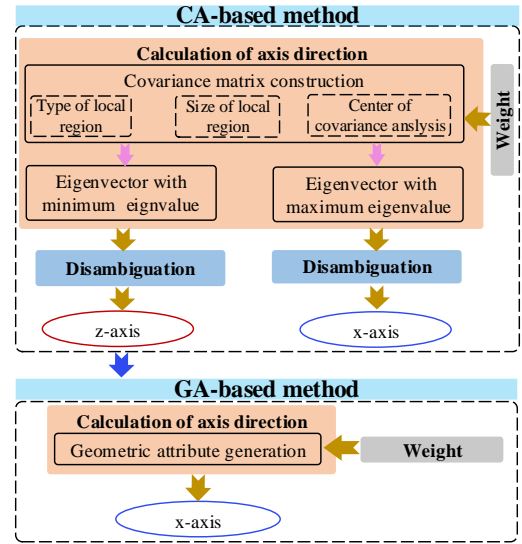

Fig. 2. General stages of the covariance analysis (CA)- and geometric attribute (GA)-based LRF methods.

TABLE I
NOTATIONS USED IN THIS ARTICLE

| Notations | Definitions |
| --- | --- |
| **p** | A keypoint used to construct LRF |
| $L(\mathbf{p}).\mathbf{x}, L(\mathbf{p}).\mathbf{y}, L(\mathbf{p}).\mathbf{z}$ | The x-, y- and z-axis of the LRF at **p** |
| $R$ | A support radius used for determining the neighbors of **p** |
| $\mathbf{N}(\mathbf{p})=\{\mathbf{p}_1, \mathbf{p}_2, ... , \mathbf{p}_k\}$ | A set of points within a distance of $R$ to **p** |
| $\mathbf{N}'(\mathbf{p})=\{\mathbf{p}_1', \mathbf{p}_2', ... , \mathbf{p}_k'\}$ | A set of points obtained by projecting the $\mathbf{N}(\mathbf{p})$ on the orthogonal plane of the z-axis. The $i$th projected point $\mathbf{p}_i'$ is calculated as: $\mathbf{p}_i'=\mathbf{pp}_i-(\mathbf{pp}_i \cdot L(\mathbf{p}).\mathbf{z}) \cdot L(\mathbf{p}).\mathbf{z}+\mathbf{p}$. |
| $r$ | Smaller support radius which is set to $R/3$ |
| $\mathbf{N}_s(\mathbf{p})=\{\mathbf{p}_{s1}, \mathbf{p}_{s2}, ... , \mathbf{p}_{sl}\}$ | A set of points within a distance of $r$ to **p** |
| $\mathbf{C}(\mathbf{Q})$ | The covariance matrix of a point set **Q** |
| $\bar{\mathbf{p}}$, $\bar{\mathbf{p}}_s$ | The barycenter of $\mathbf{N}(\mathbf{p})$ and $\mathbf{N}_s(\mathbf{p})$ respectively |
| $\mathbf{M}(\mathbf{p})=\{\mathbf{s}_1, \mathbf{s}_2, ... , \mathbf{s}_u\}$ | A mesh representation of local region within a distance of $R$ to **p**, where $\mathbf{s}_i$ denotes the $i$th triangle |
| $\mathbf{n}(\mathbf{q})$ | The normal vector of a point **q** |

*A. CA-based methods*

CA-based methods conduct the covariance analysis of local points or mesh to construct the axes of LRF. It mainly includes three steps as introduced in the above. In this process, existing methods explore the robust and repeatable LRF construction mainly from three aspects: the calculation of axis direction, the weight in constructing covariance matrix and the disambiguation of axis direction. In order to accurately validate the respective contribution of these three aspects in CA-based methods, they are described in detail as three parts.

*1) The Methods of Calculating Axis Direction*

In this part, only the main strategies for calculating the axis direction are introduced, regardless of the weights and disambiguation of the axis. The details are introduced as follows.

*a) Covariance Analysis of Local Points with Keypoint as center (CA-P-k)[29]*: This method conducts the covariance analysis on the local points $\mathbf{N}(\mathbf{p})$ with the keypoint $\mathbf{p}$ as the center. The covariance matrix is calculated as:

$$\mathbf{C}(\mathbf{N}(\mathbf{p})) = \frac{1}{|\mathbf{N}(\mathbf{p})|} \sum_{\mathbf{p}_i \in \mathbf{N}(\mathbf{p})} (\mathbf{p}_i - \mathbf{p})(\mathbf{p}_i - \mathbf{p})^T \quad (1)$$

The eigenvectors corresponding to the minimum and the maximum eigenvalues of this covariance matrix are obtained as the directions of z- and x-axis, respectively.

*b) Covariance Analysis on Local Points with Barycenter as Center (CA-P-b)[30]*: This method implements the covariance analysis on the local points $\mathbf{N}(\mathbf{p})$ with the barycenter $\bar{\mathbf{p}}$ as the center. The covariance matrix is constructed as:

$$\mathbf{C}(\mathbf{N}(\mathbf{p})) = \frac{1}{|\mathbf{N}(\mathbf{p})|} \sum_{\mathbf{p}_i \in \mathbf{N}(\mathbf{p})} (\mathbf{p}_i - \bar{\mathbf{p}})(\mathbf{p}_i - \bar{\mathbf{p}})^T \quad (2)$$

The eigenvectors corresponding to the minimum and the maximum eigenvalues of this covariance matrix are obtained as the directions of z- and x-axis, respectively.

*c) Covariance Analysis on Small Local Points with Keypoint as center (CA-sP-k)*: This method conducts the covariance analysis on the small local points $\mathbf{N}_s(\mathbf{p})$ with the keypoint $\mathbf{p}$ as the center. The purpose of reducing the scale of local region in covariance analysis is to improve the robustness to occlusion. The covariance matrix of this method is computed as:

$$\mathbf{C}(\mathbf{N}_s(\mathbf{p})) = \frac{1}{|\mathbf{N}_s(\mathbf{p})|} \sum_{\mathbf{p}_{si} \in \mathbf{N}_s(\mathbf{p})} (\mathbf{p}_{si} - \mathbf{p})(\mathbf{p}_{si} - \mathbf{p})^T \quad (3)$$

The eigenvector corresponding to the minimum eigenvalue is defined as the direction of the z-axis. It is worth noting that this strategy is only used to calculate a z-axis in existing LRF methods [9, 25] since a small local region more easily produces symmetrical feature which seriously influences the repeatability of a x-axis.

*d) Covariance Analysis on Small local Points with Barycenter as center (CA-sP-b)[9]*: This method conducts the covariance analysis on the small local points $\mathbf{N}_s(\mathbf{p})$ with the barycenter $\bar{\mathbf{p}}$ as the center. The covariance matrix is computed as:

$$\mathbf{C}(\mathbf{N}_s(\mathbf{p})) = \frac{1}{|\mathbf{N}_s(\mathbf{p})|} \sum_{\mathbf{p}_{si} \in \mathbf{N}_s(\mathbf{p})} (\mathbf{p}_{si} - \bar{\mathbf{p}}_s)(\mathbf{p}_{si} - \bar{\mathbf{p}}_s)^T \quad (4)$$

The eigenvector corresponding to the minimum eigenvalue is obtained as the direction of z-axis. Similar to CA-sP-k, this method is also only used to construct a z-axis.

*e) Covariance Analysis on local Mesh with Keypoint as center (CA-M-k)[6]*: In contrast to the above methods using the local points for covariance analysis, this method employs the local mesh in covariance analysis. Specifically, CA-M-k firstly generates the covariance analysis of each triangle in the local mesh $\mathbf{M}(\mathbf{p})$, and then aggregates the resulting matrices into an overall covariance matrix $\mathbf{C}(\mathbf{M}(\mathbf{p}))$. Specifically, the covariance analysis for a triangle $\mathbf{s}_i$ in $\mathbf{M}(\mathbf{p})$ is performed with the keypoint $\mathbf{p}$ as the center, which is calculated as:

$$\mathbf{C}(\mathbf{s}_i) = \frac{1}{12} \sum_{m=1}^{3} \sum_{n=1}^{3} (\mathbf{p}_i^m - \mathbf{p})(\mathbf{p}_i^n - \mathbf{p})^T + \frac{1}{12} \sum_{m=1}^{3} (\mathbf{p}_i^m - \mathbf{p})(\mathbf{p}_i^m - \mathbf{p})^T, (5)$$

where $\mathbf{p}_i^j$ denotes the $j$th vertice of the triangle $\mathbf{s}_i$. Based on the formula (5), the covariance analysis for each local triangle can be conducted, and then the overall covariance matrix of $\mathbf{M}(\mathbf{p})$ is obtained as:

$$\mathbf{C}(\mathbf{M}(\mathbf{p})) = \sum_{\mathbf{s}_i \in \mathbf{M}(\mathbf{p})} \mathbf{C}(\mathbf{s}_i) \quad (6)$$

The eigenvectors corresponding to the minimum and maximum eigenvalues of the covariance matrix $\mathbf{C}(\mathbf{M}(\mathbf{p}))$ are defined as the directions of z- and x-axis, respectively. The main purpose of using mesh to replace points in covariance analysis is to improve the robustness to varying mesh resolutions.

*f) Covariance Analysis on local Mesh with Barycenter as center (CA-M-b)*: This method is firstly proposed in this paper. Similar to CA-M-k, it also uses the local mesh $\mathbf{M}(\mathbf{p})$ for conducting covariance analysis. In CA-M-b, the covariance analysis for a triangle $\mathbf{s}_i$ of $\mathbf{M}(\mathbf{p})$ is performed with the barycenter $\bar{\mathbf{p}}$ as the center, which is calculated as:

$$\mathbf{C}(\mathbf{s}_i) = \frac{1}{12} \sum_{m=1}^{3} \sum_{n=1}^{3} (\mathbf{p}_i^m - \bar{\mathbf{p}})(\mathbf{p}_i^n - \bar{\mathbf{p}})^T + \frac{1}{12} \sum_{m=1}^{3} (\mathbf{p}_i^m - \bar{\mathbf{p}})(\mathbf{p}_i^m - \bar{\mathbf{p}})^T \quad (7)$$

where $\mathbf{p}_i^j$ denotes the $j$th vertice of the triangle $\mathbf{s}_i$. Based on the formula (7), the covariance analysis for each local triangle can be conducted, and the overall covariance matrix of $\mathbf{M}(\mathbf{p})$ is obtained with the formula (6). The eigenvectors corresponding to the minimum and maximum eigenvalues of the covariance matrix $\mathbf{C}(\mathbf{M}(\mathbf{p}))$ are calculated as the directions of z- and x-axis, respectively.

*g) Covariance Analysis on Projected local Points with Keypoint as center (CA-pP-k)[25]*: In this method, the local points $\mathbf{N}(\mathbf{p})$ are firstly projected along z-axis to obtain the projected local points $\mathbf{N}'(\mathbf{p})$. This process needs the z-axis to be available. Then, the covariance analysis of this method is performed on $\mathbf{N}'(\mathbf{p})$ with the keypoint $\mathbf{p}$ as center, which is calculated as:

$$\mathbf{C}(\mathbf{N}'(\mathbf{p})) = \frac{1}{|\mathbf{N}'(\mathbf{p})|} \sum_{\mathbf{p}'_i \in \mathbf{N}'(\mathbf{p})} (\mathbf{p}'_i - \mathbf{p})(\mathbf{p}'_i - \mathbf{p})^T \quad (8)$$

The eigenvector corresponding to the maximum eigenvalue of $\mathbf{C}(\mathbf{N}'(\mathbf{p}))$ is calculated as the direction of x-axis. Since the covariance analysis is performed on xy-plane, this method is only used for constructing x-axis.

*2) Weight Design Strategies*

In existing CA-based methods, weights are commonly used in computing the covariance matrix. We broadly group these weights into three categories: radial-based weight ($w_r$), area-based weight ($w_a$) and height-based weight ($w_h$). The usage of these weights in constructing covariance matrix is similar. Specifically, in the local points-based method, taken the method CA-P-k as example, the weighted covariance matrix can be calculated from the formula (1) as:

$$C(N(p)) = \frac{1}{|N(p)|} \sum_{p_i \in N(p)} w_i (p_i - p)(p_i - p)^T, \quad (9)$$

where the $w_i$ denotes the weight corresponding to a neighbor $p_i$. In the mesh-based methods, the weighted overall covariance matrix can be computed from the formula (6) as:

$$C(M(p)) = \sum_{s_i \in M(p)} w_i C(s_i), \quad (10)$$

where the $w_i$ denotes the weight corresponding to a neighbor triangle $s_i$. It is worth noting that, in the formula (9) and (10), the three weights $w_r$, $w_a$ and $w_h$ can be used alone or in combination. The details of these weights are presented as follows.

*a) Radial-based Weight ($w_r$)[29]*: This weight is associated with the radial distance which denotes the Euclidean distance between $p$ and its neighbor. The $w_r$ of the $i$th neighbor point $p_i$ is calculated as:

$$w_{ri} = (R - \|p_i - p\|_2)^2 \quad (11)$$

*b) Area-based Weight ($w_a$)[6]*: This weight is defined as the ratio of the area of each triangle to the area of all triangles in $M(p)$. The $w_a$ corresponding to the $i$th triangle $s_i$ can be calculated as:

$$w_{ai} = \frac{|(p_i^2 - p_i^1) \times (p_i^3 - p_i^1)|}{\sum_{s_j \in M(p)} |(p_j^2 - p_j^1) \times (p_j^3 - p_j^1)|} \quad (12)$$

Since this weight is related with the area of triangle, it is only used in the mesh-based methods (i.e., CA-M-k and CA-M-b).

*c) Height-based Weight ($w_h$)[25]*: This weight is relevant to the height of the neighbor point along the z-axis. Since it needs the z-axis being available, it is only used in the construction of x-axis. In existing methods, this weight has two styles: height distance [9] and height distance with a Gaussian form [25]. Without loss of generality, we consistently use the second pattern in this paper, which is calculated as:

$$w_{hi} = e^{-(\max(H) - h_i)^2 / (2\delta^2)^2}, \quad (13)$$

where $w_{hi}$ and $h_i$ denote the weight $w_h$ and the height $h_i$ of the $i$th neighbor point respectively, and $H = \{h_1, h_2, \ldots, h_k\}$ denotes the height set containing the heights of all neighbors. The $\delta$ in the formula (13) denotes the standard deviation in the Gaussian function, which is set to $\max(H)/9$ as recommended in the literature[25].

*3) Disambiguation Method*

In CA-based methods, the direction of LRF axes directly determined with the eigenvectors are ambiguous. To obtain the unambiguous axis, disambiguation is necessary. In existing methods, two common disambiguation methods are proposed. The details are given as follows.

*a) Points Mean-based Method[9]*: This method redirects the sign of x- or z-axis toward the side of the centroid of neighbors. Taking z-axis for example, the disambiguation of z-axis can be computed as:

$$L(p).z = \begin{cases} v_{min}, & \text{if } v_{min} \cdot \sum_{p_i \in N(p)} pp_i \geq 0 \\ -v_{min}, & \text{otherwise} \end{cases}, \quad (14)$$

where $v_{min}$ denotes the eigenvector corresponding to the minimum eigenvalue of a covariance matrix, and $pp_i$ represents the vector from $p$ to $p_i$.

*b) Normal Mean-based Method[25]*: This method redirects the ambiguous axis toward the side of the average direction of all neighbor normal. Taking z-axis for example, the disambiguation can be computed as:

$$L(p).z = \begin{cases} v_{min}, & \sum_{p_i \in N(p)} v_{min} \cdot n(p_i) \geq 0 \\ -v_{min} & \text{otherwise} \end{cases}, \quad (15)$$

where $n(p_i)$ represents the normal vector of a neighbor $p_i$.

*B. GA-based methods*

In contrast to the covariance analysis employed in CA-based methods, GA-based methods explore effective geometric attributes in the LRF axis construction. In existing GA-based methods, the geometric attribute is commonly designed for constructing the x-axis with the z-axis being given. Since the direction determined by the geometric attribute is definite, these GA-based methods do not need disambiguation. Similar to the CA-based methods, some of these GA-based ones can also employ weights for improving their robustness. For readability, the existing GA-based methods are decomposed into two parts (including the method of calculating axis direction and weight design strategy) for detailing as follows.

*1) The Method of Calculating Axis Direction*

*a) Geometric Attribute with the Mean of All Projected Local Points (GA-mpP)[9]*: The x-axis in this method is constructed by averaging all projected local points $N'(p)$, which is calculated as:

$$L(p).x = \frac{\sum_{p_i' \in N'(p)} pp_i'}{\left\| \sum_{p_i' \in N'(p)} pp_i' \right\|}, \quad (16)$$

where $pp_i'$ denotes the vector from $p$ to $p_i'$.

*b) Geometric Attribute with Maximum Deviation Angle (GA-mA)[28]*: In this method, only the points in the border region of $N_i$, i.e., points whose distances to the keypoint $p$ are larger than $0.85 \times R$, are considered. In these border points, the point with the largest angle between its normal and $n(p)$ is defined as the salient point. The x-axis is determined as the normalized vector from $p$ to the point obtained by projecting the salient point on the orthogonal plane of the z-axis.

*c) Geometric Attribute with Maximum Height (GA-mH)[31]*: In this method, the available points for calculating x-axis are also selected in the border region. In these border points, the point with the maximum projected height along z-axis is selected as the salient point. Based on this kind of salient point, the x-axis is finally determined by the same way as GA-mA.

*2) Weight Design Strategy*

In GA-based methods, weights are only used in GA-mpP method, which is not as common as the weights used in the CA-based methods. In GA-mpP, the radial-based weight $w_r$ and height-based weight $w_h$ can be used. These two weights have been detailed in the Sect. II-A-3). The weights used in GA-mpP is implemented by adding them in the formula (16) as:

$$L(\mathbf{p}).\mathbf{x} = \frac{\sum_{\mathbf{p}'_i \in \mathbf{N}'(\mathbf{p})} w_i \cdot \mathbf{pp}'_i}{\left\| \sum_{\mathbf{p}'_i \in \mathbf{N}'(\mathbf{p})} w_i \cdot \mathbf{pp}'_i \right\|}, \quad (17)$$

where $w_i$ denotes the weight corresponding to the $i$th projected neighbor $\mathbf{p}'_i$. It is worth noting that the weight in the formula (17) can be the $w_r$, $w_h$ or $w_r w_h$.

*C. Feature Summary of All Evaluated Methods*

Based on the details introduced in Sect. II-A and II-B, the features of all evaluated methods are summarized in Table II. These features include the type and size of local region, the center of covariance analysis and available weights, etc. Since various nuisances have different effects to z- and x-axes, the z- and x-axes are separately tested in our paper to improve the accuracy and unbiasedness of this evaluation. For readability, the z- and x-axes constructed by a method T are named as T(z) and T(x), respectively. It is worth noting that the "-" denotes the method with no such attribute.

TABLE II
THE FEATURES OF ALL EVALUATED METHODS

| General category | Axis | Type of local region | Size of local region | Center of covariance analysis | Geometrical attribute | Name | Available weight | Need disambiguation |
|---|---|---|---|---|---|---|---|---|
| CA-based | z-axis | Points | $R$ | Keypoint | - | CA-P-k(z)[29] | $w_r$ | Yes |
| | | | $R$ | Barycenter | - | CA-P-b(z)[30] | $w_r$ | Yes |
| | | | $(1/3)R$ | Keypoint | - | CA-sP-k(z) | $w_r$ | Yes |
| | | | $(1/3)R$ | Barycenter | - | CA-sP-b(z)[9] | $w_r$ | Yes |
| | | Mesh | $R$ | Keypoint | - | CA-M-k(z)[6] | $w_r, w_a, w_r w_a$ | Yes |
| | | | $R$ | Barycenter | - | CA-M-b(z) | $w_r, w_a, w_r w_a$ | Yes |
| | x-axis | Points | $R$ | Keypoint | - | CA-P-k(x)[29] | $w_r$ | Yes |
| | | | $R$ | Barycenter | - | CA-P-b(x)[30] | $w_r$ | Yes |
| | | Projected points | $R$ | Keypoint | - | CA-pP-k(x)[25] | $w_r, w_h, w_r w_h$ | Yes |
| | | Mesh | $R$ | Keypoint | - | CA-M-k(x)[6] | $w_r$ | Yes |
| | | | $R$ | Barycenter | - | CA-M-b(x) | $w_r$ | Yes |
| GA-based | x-axis | Projected points | $R$ | - | Mean of projected points | GA-mpP(x)[9] | $w_r, w_h, w_r w_h$ | No |
| | | Points | $R$ | - | Maximum angle | GA-mA(x)[28] | - | No |
| | | | $R$ | - | Maximum height | GA-mH(x)[31] | - | No |

## III. EVALUATION METHODOLOGY

*A. Datasets*

In this article, six benchmark datasets are used to comprehensively evaluate the LRF axis methods listed in Table II. These datasets include Bologna 3D retrieval (B3R) [32], UWA 3D Modeling (U3M) [33], UWA 3D object recognition (U3OR) [33], Queen's LIDAR (QuLD) [34], Kinect 3D Registration (K3R) [27] and Stereo 3D Registration (S3R) [27] datasets. Three exemplars of each dataset are presented in Fig. 3. The features of these datasets are summarized in Table III. Note that, in the registration of two point clouds, the point cloud needing to be transformed is defined as a model, and the point cloud being fixed is regarded as a scene for simplicity. In the original B3R dataset, 18 scenes are contained. They are created by adding Gaussian noises with the standard deviations of 0.1, 0.3 and 0.5mr in each randomly transformed model. To test the performance of the axis methods under the nuisances of both noise and varying mesh resolutions, a group of scenes are created by sampling the scenes, which contain 0.5mr Gaussian noise, to 1/4 of its original resolutions. Unless otherwise noted, these new created scenes are only used in this experiment. For U3M, K3R and S3R datasets, the scan pairs with the overlap ratios larger than 10% are considered for ensuring adequate overla

TABLE III
EXPERIMENTAL DATASETS AND FEATURES

| Dataset | Scenario | Challenge | Scanner | Quality | #Model | #Scene | #Matching pair |
|---|---|---|---|---|---|---|---|
| B3R | Retrieval | Gaussian noise, mesh decimation | Cyberware 3030 MS | High | 6 | 18 | 18 |
| U3M | Registration | Occlusion | Minolta vivid 910 | High | - | 75 | 433 |
| U3OR | Recognition | Occlusion, clutter | Minolta vivid 910 | High | 5 | 50 | 188 |
| QuLD | Recognition | Occlusion, clutter, real noise | NextEngine | Low | 5 | 80 | 80 |
| K3R | Registration | Occlusion, real noise, missing regions | Microsoft Kinect | Low | - | 69 | 284 |
| S3R | Registration | Occlusion, real noise, outliers, missing regions | SpaceTime Stereo | Low | - | 57 | 240 |

The main considerations for selecting these datasets are twofold. First, these datasets are obtained by various scanners

(e.g., Minolta vivid and Kinect) and applied in different scenarios (e.g., registration, recognition and retrieval), which ensures the practicability of this evaluation. Second, these datasets contain all common nuisances (including occlusion, clutter, noise, etc.), which can comprehensively evaluate the robustness of all LRF axis methods.

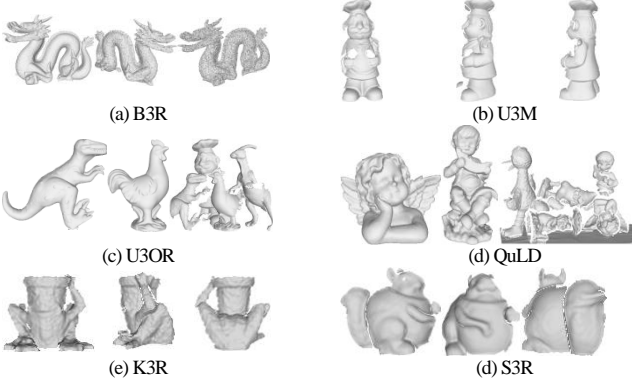

Fig. 3. Three exemplar scans respectively taken from the U3M, U3OR, B3R, QuLD, K3R and S3R datasets.

## B. Evaluation criteria

### 1) Repeatability Criterion

Two axes (e.g., $\mathbf{v}(\mathbf{p}^m)$ and $\mathbf{v}(\mathbf{q}^s)$) which are calculated on two corresponding keypoints ($\mathbf{p}^m$ and $\mathbf{q}^s$) between the model and scene are regarded as "repeatable" if the angle error $E$ between them is less than a small threshold. In this paper, the small threshold is set to 5° and the angle error $E$ between any two axes $\mathbf{v}(\mathbf{p}^m)$ and $\mathbf{v}(\mathbf{q}^s)$ is easily computed as:

$$E = \arccos\left(\frac{\mathbf{v}(\mathbf{p}^m)\tilde{\mathbf{v}}(\mathbf{p}^s)^{\mathrm{T}}}{\|\mathbf{p}^m\|\|\mathbf{p}^s\|}\right), \quad (18)$$

where $\tilde{\mathbf{v}}(\mathbf{p}^m) = \mathbf{R}_{GT}\mathbf{v}(\mathbf{p}^m)$ with $\mathbf{R}_{GT}$ being the ground truth rotation between the model and scene. Based on this criterion, the statistic of repeatability is implemented as follows. First, 1000 keypoints are randomly selected on scene, and the 1000 corresponding keypoints on model/models are determined by the ground truth transformation which is provided by the publishers or obtained by manual alignment and the iterative closest point (ICP) algorithm [35]. Second, the methods listed in Table II are used to generate LRF axes on each keypoint, and the percentage of "repeatable" axes are counted for evaluating the repeatability of these axes.

This criterion is used for evaluating the repeatability and robustness of all tested axis methods on the six experimental datasets.

### 2) Feature Matching Criterion

LRF or its z-axis (called LRA) is the basis for generating most local feature descriptors. The repeatability of the LRF axes seriously influences the performance of these descriptors. In this paper, the influence of different LRF axes to the matching performance of descriptor is evaluated. The recall vs 1-precision curve (RPC) [8, 10] is a commonly used criterion for feature matching evaluation. The detail of generating RPC in feature matching evaluation can reference articles [9, 10]. To compactly and quantitatively present the performance of feature descriptors, the area under precision vs recall curve, defined as $AUC_{pr}$, is used in this paper. $AUC_{pr}$ is a simple and aggregated metric to measure how an algorithm performs over the whole precision-recall space [10]. In an ideal case, the $AUC_{pr}$ equals to 1 since the precision is always 1 for any recall.

### 3) Point Cloud Registration Criterion

LRF or LRA (i.e., the z axis of LRF) is an important tool to design transformation estimation methods in aligning two point clouds. The influence of different LRF axes to the registration performance of these LRF- or LRA-based transformation estimation methods are also tested in this paper. The rotation error $\varepsilon_r$ between the estimated rotation matrix $\mathbf{R}_E$ and the ground truth one $\mathbf{R}_{GT}$ and the translation error $\varepsilon_t$ between the estimated translation vector $\mathbf{t}_E$ and the ground truth one $\mathbf{t}_{GT}$ are calculated on each model-scene pair. Like [15], the rotation error $\varepsilon_r$ and the translation error $\varepsilon_t$ are calculated as:

$$\begin{cases} \varepsilon_r = \arccos\left(\frac{trace(\mathbf{R}_{GT}(\mathbf{R}_E)^{-1})-1}{2}\right)\frac{180}{\pi} \\ \varepsilon_t = \frac{\|\mathbf{t}_{GT}-\mathbf{t}_E+(\mathbf{R}_{GT}\bar{\mathbf{P}}^m-\mathbf{R}_E\bar{\mathbf{P}}^m)\|}{\mathrm{mr}} \end{cases}, \quad (19)$$

where mr denotes the average mesh resolution and $\bar{\mathbf{P}}^m$ denotes the center of model. For an estimated transformation ($\mathbf{R}_E$, $\mathbf{t}_E$), if $\varepsilon_r$ below 5° and $\varepsilon_t$ below 5, it is deemed as a correct transformation, otherwise it is regarded as a false one.

## C. Evaluation Items

### 1) Support Radius

The support radius is an important parameter for constructing LRF. Different axis methods or experimental datasets usually have different optimal support radii. To explore the influence of different support radii to these axis methods as well as set the optimal support radii for these axis methods on different datasets, we set the support radii increasing from 5 to 30mr with an interval of 5mr for testing all the axis methods. Based on the results, the optimal support radii for these methods on different datasets can be determined, and they will be used in all the following test items for an unbiased evaluation.

### 2) Weight

To improve the robustness of LRF methods, various weights are often used, as summarized in Sect. II. The merits and demerits of these weights are important for the LRF construction. In this test, the weights are comprehensively tested on all the six experimental datasets. Based on the results, the optimal weights for these methods are set, and used in the following test items for an unbiased evaluation.

### 3) Disambiguation

Disambiguation is commonly necessary in CA-based LRF construction. The technique of disambiguation mainly includes points mean-based and normal mean-based methods, as presented in Sect. II. In this test, the merits and demerits of the two disambiguation methods are evaluated on all the six datasets, and the optimal disambiguation methods are selected and used in all the following test items.

*4) Influence of the z-axes to the x-axes*

Some x-axis methods are constructed on the basis of z-axes. These z-axis-dependent x-axis methods include CA-pP-k(x), GA-mpP(x), GA-mA(x) and GA-mH(x) as listed in Table II. To evaluate the influence of different z-axes to the x-axis methods and set the optimal z-axis for these x-axis methods in the following test items, the x-axis methods combined with different z-axis methods are tested.

*5) Repeatability on Different Datasets*

To comprehensively evaluate the repeatability of all the fourteen axis methods, these axis methods are tested on the six experimental datasets. These datasets have different nuisances and scenarios, and are acquired by various scanners. They totally contain 1243 model-scene pairs. Thus, it can comprehensively evaluate the repeatability of the axis methods.

*6) Robustness*

The robustness of these LRF axes is evaluated by testing the repeatability of them under different levels of various nuisances. In this paper, eight main nuisances (including Gaussian noise, varying mesh resolutions, distance to boundary, keypoint location error, occlusion, clutter and partial overlap) are considered.

*a) Gaussian Noise*: This is tested on the B3R dataset. Since this dataset does not contain other nuisances, it ensures the accuracy of this evaluation. To comprehensively test the robustness to noise, the group of scenes are generated by adding the Gaussian noise with the standard deviation increased from 0.0 to 1.0mr with an interval of 0.2mr in each randomly transformed B3R model, where mr denotes the average mesh resolution. An illustration of a point cloud with 1.0mr Gaussian noise is exhibited in Fig. 4(b).

*b) Varying Mesh Resolutions*: Without lose of generality, we also use the B3R dataset to test this kind of robustness. In this test, the group of scenes are generated by sampling each randomly transformed model from 1/1 (i.e., the original resolution) to 1/32 of its original mesh resolution with an interval of 1/10. An illustration of a point cloud with 1/32 mesh decimation is shown in Fig. 4(c).

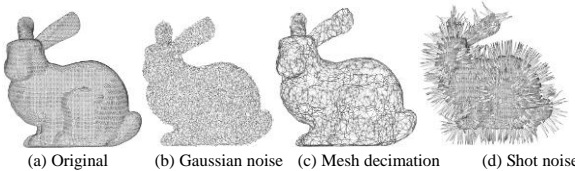

(a) Original    (b) Gaussian noise    (c) Mesh decimation    (d) Shot noise

Fig. 4. An illustration of a shape with different nuisances. From left to right: a clean 3D shape, a shape with 1.0mr Gaussian noise, 1/32 mesh decimation and 3% shot noise.

*c) Shot Noise*: The shot noise used in this test is generated by the displacement of a given point along normal direction. Since shot noise usually lies in the view line of a given point, this method can well simulate the real noise [36]. Based on this process, we produce six levels of shot noise with the outlier ratios of 0.1%, 0.3%, 0.5%, 0.8%, 1%, 3% and 5%. The displacement amplitude of these outliers along normal direction is set to $R$, where $R$ is set to 20mr in this test. An illustration of a point cloud with 3% shot noise is presented in Fig. 4(d).

*d) Distance to Boundary*: Different distances to boundary commonly leads to different intensities of occlusion which is a main nuisance for influencing the performance of LRFs. For comprehensively testing this item, five levels of the distance range to boundary are set. They are $(0, 0.2R]$, $(0.2R, 0.4R]$, $(0.4R, 0.6R]$, $(0.6R, 0.8R]$, $(0.8R, 1.0R]$ and larger than $1.0R$, where $R$ is set to 20mr in this test. Without lose of generality, this evaluation is only tested on the U3OR dataset since this dataset has more boundaries than the others. An illustration for the six levels of distance range is shown in Fig. 5.

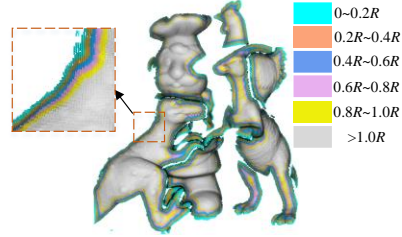

Fig. 5. An illustration of six areas with different distances to boundary.

*e) Keypoint Localization Error*: In practical application, keypoint localization error is inevitable. Thus, the robustness to keypoint localization error is also evaluated for these LRF axes. This evaluation is tested on the U3OR dataset. The keypoint localization errors is set to six levels, including the localization errors of $0.0R$, $0.2R$, $0.4R$, $0.6R$, $0.8R$ and $1.0R$, where $R$ is set to 20mr.

*f) Occlusion*: It is a common nuisance in object recognition applications. The degree of this nuisance can be calculated as:

$$\text{occlusion} = 1 - \frac{\text{model surface area in scene}}{\text{total model surface area}} \quad (20)$$

In this test, the U3OR, a popular object recognition dataset, is used. We divide this dataset into seven groups according to the degree of occlusion in less than 65%, [65%, 70%), [70%, 75%), [75%, 80%), [80%, 85%), [85%, 90%) and [90%, 100%].

*g) Clutter*: It is also a common nuisance in object recognition applications. The degree of clutter can be calculated as [10]:

$$\text{clutter} = 1 - \frac{\text{model surface area in scene}}{\text{total surface area of scene}} \quad (21)$$

In this test, the U3OR dataset also be used. According to the degree of clutter, we divide this dataset into eight groups, which have the degree of clutter in less than 60%, [60%, 65%), [65%, 70%), [70%, 75%), [75%, 80%), [80%, 85%), [85%, 90%) and [90%, 100%].

*h) Partial Overlap:* In practical, the obtained point clouds are commonly partial overlap, and the overlap ratios of these point clouds are different. To verify the influence of varying overlap ratios on these axis methods, their repeatability with respect to different overlap ratios are tested. The overlap ratio [37] between two point clouds is calculated as:

$$\text{overlap} = \frac{\text{\# points in the overlap area}}{\min(\text{\# model points, \# scene points})} \quad (22)$$

In this test, we use the U3M dataset, and divide the point cloud pairs in this dataset into eight groups according to their overlap ratios in the range of less than 30%, [30%, 40%), [40%, 50%), [50%, 60%), [60%, 70%), [70%, 80%), [80%, 90%) and [90%, 100%].

*7) Feature Matching Performance*

A LRF and the z-axis in a LRF (named as LRA) are the basis for constructing most local feature descriptors [10, 20, 21]. The performance of the LRF- or LRA-based descriptors are seriously influenced by the repeatability of the LRF axes [8, 19, 20, 38]. This item evaluates the influence of different axis methods to the feature matching performance of these LRF- and LRA-based descriptors. Specifically, three LRA-based (including Spin Image [39], LFSH [40] and MaSH [24]) and three LRF-based (including RoPS [6], SHOT [41] and TOLDI [9]) descriptors are selected in this evaluation. All the z-axis methods listed in Table II are used to generate the three LRA-based descriptors, and all the x-axis methods listed in Table II combining with the CA-sP-b(z) are used to construct the three LRF-based descriptors. Without lose of generality, this evaluation is tested on the U3M dataset.

*8) Transformation Estimation Performance*

The LRF and LRA are the important tools for estimating the transformation between two point clouds under the point-to-point correspondences being given [15]. The repeatability of LRF and LRA are seriously relevant to the accuracy of these LRF- and LRA-based transformation estimation methods. This item evaluates the influence of different axis methods to the accuracy of the LRF- and LRA-based transformation estimation methods. Specifically, two LRA-based (including 2P-RANSAC [24] and CG-SAC [42]) and three LRF-based (including CCV [6], 1P-RANSAC [43] and LRF-MCS [26]) transformation estimation methods are considered. Without lose of generality, this evaluation is tested on the U3M dataset.

## IV. EXPERIMENTAL RESULTS

This section gives the results of the test terms introduced in Sect. III-C together with necessary discussions and explanations.

### A. Support Radius

The repeatability of the fourteen axis methods with respect to different support radii on the six experimental datasets are shown in Fig. 6. Several observations can be found as follows.

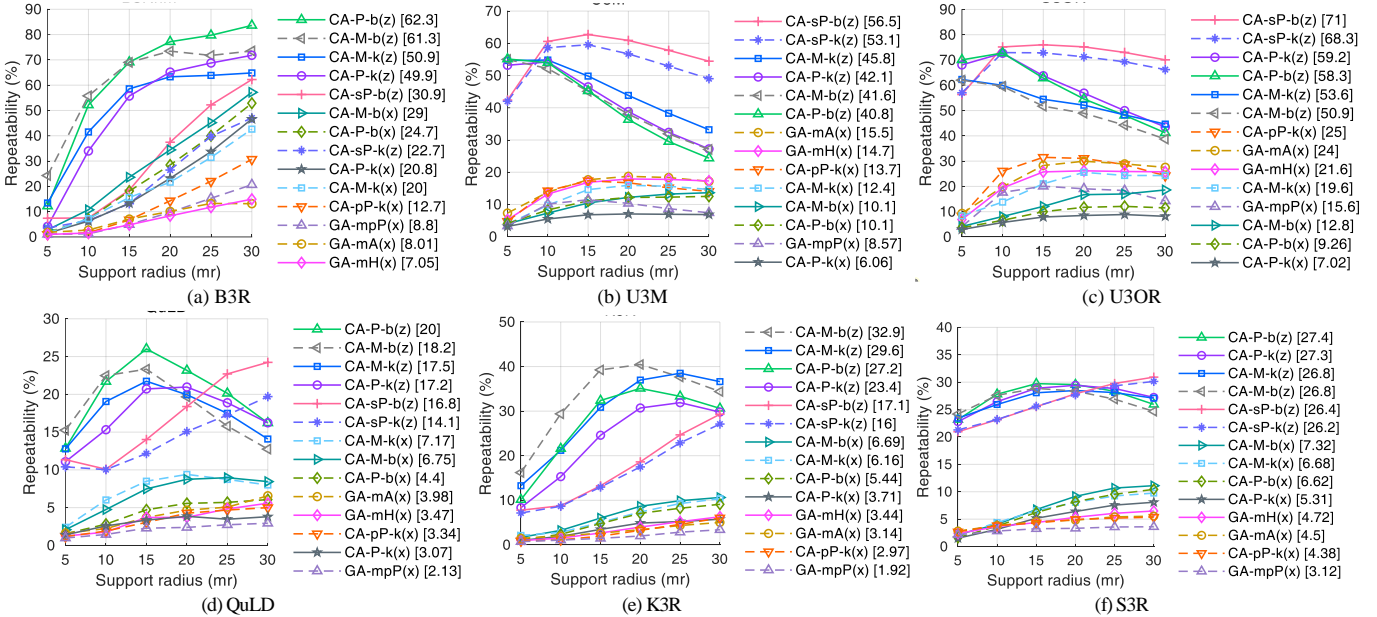

Fig. 6. The repeatability of the fourteen axis methods with respect to different support radii tested on the six experimental datasets.

*First*, on the B3R dataset which contains the nuisances of noise and varying mesh resolutions, the performance of all axis methods is improved along with the increase of support radius. The reason is that this dataset only contains noise and varying mesh resolutions, and a large scale of local region is commonly robust to noise and varying mesh resolutions. *Second*, on the U3M and U3OR datasets which mainly contain the nuisance of occlusion, the performance of most x-axis methods (e.g., CA-P-k(x), CA-M-k(x) and CA-M-b(x)) improves as the support radius increased from 5 to 10 or 20mr, and then almost not improves as the support radius further increases. The performance of most z-axis methods (e.g., CA-sP-b(z) and CA-sP-k(z)) improves as the support radius increased from 5 to 10 or 15mr, and then drops as the support radius further increases. The two mesh-based z-axis methods (i.e., CA-M-k(z) and CA-M-b(z)) present a worse performance along with the increase of support radius. It can verify that using mesh to construct z-axis is very sensitive to occlusion. *Third*, on the three low-quality datasets (i.e., QuLD, K3R and S3R) which contain strong noise and occlusion, the performance of most x-axis and two z-axis methods (i.e., CA-sP-b(z) and CA-sP-k(z)) gradually improves as the support radius increases, and the performance of the remaining z-axis methods firstly improves and then drops along with the increase of the support radius. It indicates that the z-axis methods except CA-sP-b(z) and CA-sP-k(z) are more vulnerable to occlusion than the x-axis methods.

Based on the result of this evaluation as well as the tradeoff between efficiency and repeatability, the support radii for the fourteen axis methods tested on the six experimental datasets are set in Table IV. In the below, unless otherwise noted, these support radius sets are used in all tests.

TABLE IV
THE SUPPORT RADIUS SETTINGS FOR THE FOURTEEN AXIS METHODS ON THE SIX EXPERIMENTAL DATASETS

|  | B3R | U3M | U3OR | QuLD | K3R | S3R |
|---|---|---|---|---|---|---|
| CA-P-k(z) | 20mr | 10mr | 10mr | 20mr | 20mr | 20mr |
| CA-P-b(z) | 20mr | 10mr | 10mr | 15mr | 20mr | 15mr |
| CA-sP-k(z) | 20mr | 15mr | 10mr | 20mr | 20mr | 20mr |
| CA-sP-b(z) | 20mr | 15mr | 15mr | 20mr | 20mr | 20mr |
| CA-M-k(z) | 20mr | 10mr | 5mr | 15mr | 20mr | 20mr |
| CA-M-b(z) | 20mr | 5mr | 5mr | 15mr | 20mr | 15mr |
| CA-P-k(x) | 20mr | 20mr | 20mr | 20mr | 20mr | 20mr |
| CA-P-b(x) | 20mr | 20mr | 20mr | 20mr | 20mr | 20mr |
| CA-pP-k(x) | 20mr | 15mr | 15mr | 20mr | 20mr | 20mr |
| CA-M-k(x) | 20mr | 20mr | 20mr | 20mr | 20mr | 20mr |
| CA-M-b(x) | 20mr | 20mr | 20mr | 20mr | 20mr | 20mr |
| GA-mpP(x) | 20mr | 15mr | 15mr | 20mr | 20mr | 20mr |
| GA-mA(x) | 20mr | 20mr | 20mr | 20mr | 20mr | 20mr |
| GA-mH(x) | 20mr | 20mr | 20mr | 20mr | 20mr | 20mr |

## B. Weight

Table V presents the repeatability of the twelve axis methods tested on the six datasets with respect to different weights. Note that the $w_0$ denotes the weight strategy of setting the same value for each neighbor. Several observations are presented as follows. *First*, the $w_0$ is more appropriate for these barycenter-based methods (e.g., CA-P-b(z) and CA-M-b(z)), and the $w_r$ is more appropriate for the keypoint-based methods (e.g., CA-P-k(z) and CA-M-k(z)). *Second*, for the mesh-based axis methods (i.e., CA-M-k(z), CA-M-b(z), CA-M-k(x) and CA-M-b(x)), the $w_a$ is an effective weight for improving their performance. *Third*, the $w_h$ is a very effective weight strategy for CA-pP-k(x) and GA-mpP(x), improving their performance significantly.

Based on the above observations, the superior weights for constructing the twelve axis methods are selected, and summarized in Table VI. In the following tests, the weight set of these twelve LRF axes consistently employs these weight values.

TABLE V
THE REPEATABILITY FOR THE TWELVE AXIS METHODS AGAINST DIFFERENT WEIGHTS TESTED ON THE SIX EXPERIMENTAL DATASETS. THE "$w_0$" DENOTES THE WEIGHT STRATEGY OF SETTING THE SAME VALUE FOR EACH NEIGHBOR, AND THE OPTIMAL PERFORMANCE IS SHOWN IN BOLD FACE

|  | B3R | U3M | U3OR | QuLD | K3R | S3R | Average |
|---|---|---|---|---|---|---|---|
| CA-P-k(z) | $w_0$:61, $w_r$:**64** | $w_0$:47, $w_r$:**54** | $w_0$:68, $w_r$:**73** | $w_0$:15, $w_r$:**20** | $w_0$:29, $w_r$:**31** | $w_0$:26, $w_r$:**29** | $w_0$:41, $w_r$:**45** |
| CA-P-b(z) | $w_0$:**77**, $w_r$:63 | $w_0$:**54**, $w_r$:51 | $w_0$:**74**, $w_r$:71 | $w_0$:**25**, $w_r$:18 | $w_0$:**35**, $w_r$:28 | $w_0$:**30**, $w_r$:27 | $w_0$:**49**, $w_r$:43 |
| CA-sP-k(z) | $w_0$:26, $w_r$:26 | $w_0$:59, $w_r$:**60** | $w_0$:71, $w_r$:**72** | $w_0$:15, $w_r$:15 | $w_0$:**19**, $w_r$:18 | $w_0$:28, $w_r$:28 | $w_0$:36, $w_r$:**37** |
| CA-sP-b(z) | $w_0$:**37**, $w_r$:**37** | $w_0$:**63**, $w_r$:**63** | $w_0$:**78**, $w_r$:**78** | $w_0$:**19**, $w_r$:**19** | $w_0$:**19**, $w_r$:**19** | $w_0$:**28**, $w_r$:**28** | $w_0$:**41**, $w_r$:40 |
| CA-M-k(z) | $w_0$:56, $w_r$:57, $w_a$:60, $w_rw_a$:**61** | $w_0$:43, $w_r$:49, $w_a$:49, $w_rw_a$:**55** | $w_0$:59, $w_r$:60, $w_a$:**61**, $w_rw_a$:**61** | $w_0$:15, $w_r$:17, $w_a$:19, $w_rw_a$:**22** | $w_0$:32, $w_r$:33, $w_a$:**38**, $w_rw_a$:**38** | $w_0$:26, $w_r$:**27**, $w_a$:26, $w_rw_a$:**27** | $w_0$:38, $w_r$:41, $w_a$:42, $w_rw_a$:**44** |
| CA-M-b(z) | $w_0$:71, $w_r$:56, $w_a$:**72**, $w_rw_a$:58 | $w_0$:53, $w_r$:52, $w_a$:**55**, $w_rw_a$:52 | $w_0$:61, $w_r$:60, $w_a$:**62**, $w_rw_a$:60 | $w_0$:**21**, $w_r$:16, $w_a$:**22**, $w_rw_a$:19 | $w_0$:37, $w_r$:31, $w_a$:**41**, $w_rw_a$:32 | $w_0$:28, $w_r$:27, $w_a$:**29**, $w_rw_a$:26 | $w_0$:45, $w_r$:40, $w_a$:**47**, $w_rw_a$:41 |
| CA-P-k(x) | $w_0$:**24**, $w_r$:19 | $w_0$:7, $w_r$:**8** | $w_0$:8, $w_r$:**11** | $w_0$:**4** $w_r$:**4** | $w_0$:**5** $w_r$:4 | $w_0$:**7**, $w_r$:6 | $w_0$:**9**, $w_r$:**9** |
| CA-P-b(x) | $w_0$:**29**, $w_r$:20 | $w_0$:**12**, $w_r$:6 | $w_0$:**11**, $w_r$:10 | $w_0$:**6**, $w_r$:3 | $w_0$:**7**, $w_r$:3 | $w_0$:**8**, $w_r$:4 | $w_0$:**12**, $w_r$:8 |
| CA-pP-k(x) | $w_0$:**16**, $w_r$:13, $w_h$:14, $w_rw_h$:15 | $w_0$:8, $w_r$:9, $w_h$:16, $w_rw_h$:**17** | $w_0$:10, $w_r$:12, $w_h$:28, $w_rw_h$:**32** | $w_0$:**4**, $w_r$:**4**, $w_h$:**4**, $w_rw_h$:**4** | $w_0$:3, $w_r$:3, $w_h$:3, $w_rw_h$:**3** | $w_0$:**5**, $w_r$:**5**, $w_h$:**5**, $w_rw_h$:**5** | $w_0$:8, $w_r$:7, $w_h$:12, $w_rw_h$:**13** |
| CA-M-k(x) | $w_0$:21, $w_r$:15, $w_a$:**29**, $w_rw_a$:24 | $w_0$:9, $w_r$:10, $w_a$:12, $w_rw_a$:**16** | $w_0$:9, $w_r$:10, $w_a$:19, $w_rw_a$:**26** | $w_0$:4, $w_r$:4, $w_a$:7, $w_rw_a$:**10** | $w_0$:7, $w_r$:5, $w_a$:**9**, $w_rw_a$:8 | $w_0$:**7**, $w_r$:**7**, $w_a$:5, $w_rw_a$:4 | $w_0$:9, $w_r$:9, $w_a$:13, $w_rw_a$:**14** |
| CA-M-b(x) | $w_0$:27, $w_r$:15, $w_a$:**35**, $w_rw_a$:21 | $w_0$:**17**, $w_r$:8, $w_a$:12, $w_rw_a$:4 | $w_0$:12, $w_r$:11, $w_a$:**16**, $w_rw_a$:9 | $w_0$:7, $w_r$:4, $w_a$:**9**, $w_rw_a$:5 | $w_0$:**11**, $w_r$:4, $w_a$:8, $w_rw_a$:3 | $w_0$:**10**, $w_r$:5, $w_a$:6, $w_rw_a$:3 | $w_0$:14, $w_r$:9, $w_a$:**15**, $w_rw_a$:8 |
| GA-mpP(x) | $w_0$:9, $w_r$:5, $w_h$:**11**, $w_rw_h$:10 | $w_0$:4, $w_r$:5, $w_h$:11, $w_rw_h$:**12** | $w_0$:6, $w_r$:7, $w_h$:**20**, $w_rw_h$:**20** | $w_0$:2, $w_r$:2, $w_h$:2, $w_rw_h$:**3** | $w_0$:**2**, $w_r$:1, $w_h$:**2**, $w_rw_h$:**2** | $w_0$:2, $w_r$:2, $w_h$:**3**, $w_rw_h$:**3** | $w_0$:4, $w_r$:4, $w_h$:**8**, $w_rw_h$:**8** |

TABLE VI
WEIGHT SETTING FOR THE TWELVE AXIS METHODS

|  | CA-P-k(z) | CA-P-b(z) | CA-sP-k(z) | CA-sP-b(z) | CA-M-k(z) | CA-M-b(z) | CA-P-k(x) | CA-P-b(x) | CA-pP-k(x) | CA-M-k(x) | CA-M-b(x) | GA-mpP(x) |
|---|---|---|---|---|---|---|---|---|---|---|---|---|
| Weight | $w_r$ | $w_0$ | $w_r$ | $w_0$ | $w_rw_a$ | $w_a$ | $w_0$ | $w_0$ | $w_rw_h$ | $w_rw_a$ | $w_a$ | $w_rw_h$ |

## C. Disambiguation

Table VII exhibits the repeatability of the eleven CA-based axis methods tested on the six datasets with respect to different disambiguation methods. Three observations can be made from the result. *First*, the normal mean-based disambiguation is better than the points mean-based one on the U3M and U3OR datasets, while their performance is comparable on the remaining datasets. It indicates that the normal mean-based disambiguation has stronger robustness to occlusion than the points mean-based one. *Second*, the normal mean-based disambiguation is obviously superior to the points mean-based one for some LRF axes when tested on the U3M, U3OR and QuLD datasets. Specifically, the CA-M-k(z) and CA-M-b(z) combined with the normal mean-based disambiguation are obviously superior to that combined with the points mean-based one on the U3M, U3OR and QuLD datasets, and the repeatability of CA-pP-k(x) has an obvious improvement from that combined with the points mean-based method to that combined with the normal mean-based one on the U3M and U3OR datasets. Based on the above observation, we consistently use the normal mean-based disambiguation for all the eleven CA-based axis methods.

TABLE VII
THE REPEATABILITY FOR THE TEN AXIS METHODS WITH RESPECT TO DIFFERENT DISAMBIGUATION METHODS TESTED ON THE SIX EXPERIMENTAL DATASETS. THE "p" AND "n" IN THIS TABLE DENOTE THE POINTS MEAN- AND NORMAL MEAN-BASED METHODS RESPECTIVELY, AND THE OPTIMAL PERFORMANCE IS SHOWN IN BOLD FACE

|  | B3R | U3M | U3OR | QuLD | K3R | S3R | Average |
|---|---|---|---|---|---|---|---|
| CA-P-k(z) | **p:65**, **n:65** | p:54, **n:56** | p:72, **n:73** | p:21, **n:22** | p:31, **n:32** | p:29, **n:30** | p:45, **n:46** |
| CA-P-b(z) | **p:78**, n:76 | p:54, **n:56** | p:72, **n:73** | **p:26**, **n:27** | p:35, **n:37** | p:30, **n:32** | p:49, **n:50** |
| CA-sP-k(z) | **p:26**, n:25 | p:59, **n:61** | p:72, **n:73** | **p:16**, **n:16** | p:18, **n:19** | **p:28**, **n:28** | p:36, **n:37** |
| CA-sP-b(z) | **p:37**, **n:37** | p:63, **n:65** | p:77, **n:79** | p:19, **n:20** | **p:19**, **n:19** | **p:28**, **n:28** | p:40, **n:41** |
| CA-M-k(z) | **p:62**, n:61 | p:55, **n:61** | p:61, **n:79** | p:22, **n:28** | **p:37**, **n:37** | p:29, **n:30** | p:44, **n:49** |
| CA-M-b(z) | **p:71**, **n:71** | p:55, **n:65** | p:62, **n:79** | p:23, **n:29** | **p:40**, n:39 | p:29, **n:30** | p:47, **n:52** |
| CA-P-k(x) | **p:25**, n:24 | p:7, **n:9** | p:9, **n:12** | **p:4**, **n:4** | **p:5**, **n:5** | p:6, **n:7** | p:9, **n:10** |
| CA-P-b(x) | **p:29**, **n:29** | p:12, **n:17** | p:12, **n:15** | **p:6**, **n:6** | **p:7**, **n:7** | p:8, **n:10** | p:12, **n:14** |
| CA-pP-k(x) | **p:14**, **n:14** | p:17, **n:27** | p:31, **n:44** | p:4, **n:5** | p:3, **n:4** | p:5, **n:7** | p:13, **n:17** |
| CA-M-k(x) | **p:24**, n:22 | **p:16**, **n:16** | p:24, **n:32** | **p:9**, **n:9** | **p:8**, n:6 | **p:8**, **n:8** | p:15, **n:16** |
| CA-M-b(x) | **p:35**, n:31 | p:12, **n:13** | p:16, **n:20** | **p:9**, n:8 | **p:9**, n:6 | **p:9**, n:8 | **p:15**, n:14 |

*D. Influence of the z-axes to the x-axes*

The repeatability of the four z-axis-dependent x-axis methods (i.e., CA-pP-k(x), GA-mpP(x), GA-mA(x) and GA-mH(x)) with respect to different z-axes is shown in Table VIII. Overall, these x-axis methods are positively relevant to the z-axis methods in terms of repeatability. On B3R dataset, these x-axis methods combined with CA-sP-k(z) and CA-sP-b(z) are obviously inferior to that combined with other z-axis methods. The reason is that CA-sP-k(z) and CA-sP-b(z) is constructed by a smaller local region, which is vulnerable to noise and varying mesh resolutions. On other datasets, the performance gaps of the x-axis methods combined with the six z-axes are small, which is because the z-axes have smaller performance gaps on U3M and U3OR datasets and the performance of the x-axes are too inferior to obtain obvious difference on QuLD, K3R and S3R datasets. Based on this result, to unbiased evaluation of all x-axis methods, we hereinafter use CA-P-b(z) on B3R and S3R datasets and CA-M-b(z) on the remaining datasets for constructing the four z-axis-dependent x-axes.

TABLE VIII
THE REPEATABILITY OF FOUR z-AXIS-DEPENDENT x-AXIS METHODS WITH RESPECT TO DIFFERENT z-AXIS METHODS ON THE SIX EXPERIMENTAL DATASETS. THE "CA-" AND "(z)" ARE OMITTED FROM THE NAMES OF THESE z-AXIS METHODS FOR SIMPLICITY, AND THE OPTIMAL PERFORMANCE IS SHOWN IN BOLD FACE.

|  | B3R | U3M | U3OR | QuLD | K3R | S3R |
|---|---|---|---|---|---|---|
| CA-pP-k(x) | P-k:23, **P-b:26**, sP-k:12, sP-b:15, M-k:24, **M-b:26** | P-k:25, P-b:24, sP-k:26, **sP-b:27**, **M-k:27**, **M-b:27** | P-k:42, P-b:43, sP-k:39, sP-b:43, M-k:43, **M-b:44** | P-k:5, P-b:6, sP-k:4, sP-b:5, **M-k:7**, **M-b:7** | P-k:6, P-b:6, sP-k:4, sP-b:4, M-k:6, **M-b:7** | P-k:7, **P-b:8**, sP-k:7, sP-b:7, M-k:7, **M-b:8** |
| GA-mpP(x) | **P-k:22**, P-b:18, sP-k:9, sP-b:11, M-k:17, M-b:16 | P-k:12, P-b:9, **sP-k:11**, sP-b:12, M-k:10, **M-b:11** | **P-k:23**, P-b:18, sP-k:18, sP-b:21, M-k:20, M-b:22 | P-k:3, P-b:3, sP-k:2, sP-b:3, **M-k:4**, **M-b:4** | P-k:2, P-b:2, sP-k:2, sP-b:2, M-k:2, **M-b:2** | **P-k:3**, **P-b:3**, **sP-k:3**, **sP-b:3**, **M-k:3**, M-b:2 |
| GA-mA(x) | P-k:15, **P-b:19**, sP-k:8, sP-b:11, M-k:16, M-b:18 | P-k:17, P-b:17, **sP-k:18**, **sP-b:18**, **M-k:18**, **M-b:18** | P-k:30, P-b:30, sP-k:29, **sP-b:31**, **M-k:31**, **M-b:31** | P-k:5, **P-b:6**, sP-k:4, sP-b:5, **M-k:6**, **M-b:6** | **P-k:5**, **P-b:5**, sP-k:3, sP-b:3, M-k:5, **M-b:5** | **P-k:5**, **P-b:5**, **sP-k:5**, **sP-b:5**, **M-k:5**, **M-b:5** |
| GA-mH(x) | P-k:16, **P-b:19**, sP-k:7, sP-b:9, M-k:15, M-b:17 | P-k:17, P-b:17, sP-k:17, **sP-b:18**, **M-k:18**, **M-b:18** | P-k:25, **P-b:26**, sP-k:24, **sP-b:26**, **M-k:26**, **M-b:26** | P-k:5, P-b:5, sP-k:4, sP-b:4, M-k:6, **M-b:7** | P-k:5, **P-b:6**, sP-k:4, sP-b:4, **M-k:6**, **M-b:6** | **P-k:6**, **P-b:6**, sP-k:5, sP-b:5, **M-k:6**, **M-b:6** |
| Average | P-k:19, **P-b:20**, sP-k:9, sP-b:11, M-k:18, M-b:19 | P-k:18, P-b:17, sP-k:18, **sP-b:19**, M-k:18, **M-b:19** | P-k:30, P-b:29, sP-k:28, sP-b:30, M-k:30, **M-b:31** | P-k:5, P-b:5, sP-k:4, sP-b:4, **M-k:6**, **M-b:6** | **P-k:5**, **P-b:5**, sP-k:3, sP-b:3, **M-k:5**, **M-b:5** | P-k:5, **P-b:6**, sP-k:5, sP-b:5, M-k:5, M-b:5 |

*E. Repeatability Performance on Different Datasets*

Based on the settings of support radius, weight, disambiguation and the appropriate z-axes for constructing the four z-axis-dependent x-axes in Sect. IV-A, IV-B, IV-C and IV-D respectively, all the fourteen methods listed in Table II have been equipped with their best parameters for an unbiased evaluation in the following tests.

The repeatability performance of all fourteen methods tested on the six experimental datasets is presented in Fig. 7. Overall, the x-axes are obviously inferior to the z-axes. This is because the x-axes are more vulnerable to various nuisances (e.g., symmetrical surface, noise and varying mesh resolutions) than the z-axes.

In terms of all z-axis methods, several observations can be made. *First*, the barypoint-based z-axes (i.e., CA-P-b(z), CA-sP-b(z) and CA-M-b(z)) are commonly superior to the keypoint-based counterparts (i.e., CA-P-k(z), CA-sP-k(z) and CA-M-k(z)), which can clearly verify that the barycenter is obviously superior to the keypoint in the z-axis construction. This is because the barycenter can relieve the affect of noise and varying mesh resolutions, as well as make the center in covariance analysis far from the boundary of object. *Second*, the mesh-based z-axes (i.e., CA-M-b(z) and CA-M-k(z)) present superior performance on all the six experimental datasets. The reason is that the mesh-based methods use all information on a local region rather than only the information of vertices as used in the other methods, improving the robustness to various nuisances, particularly noise and varying mesh resolutions. *Third*, in terms of the points-based z-axes, the z-axis methods with small scale of local region (i.e., CA-sP-k(z) and CA-sP-b(z)) achieve superior performance on U3M and U3OR datasets, while exhibit an inferior performance on other datasets. In contrast, the z-axes with normal scale of local region present the opposite performance, namely having inferior repeatability on the U3M and U3OR datasets and superior repeatability on other datasets. This is because the small scale of local region is robust to occlusion, while susceptible to noise, outliers and varying mesh resolutions.

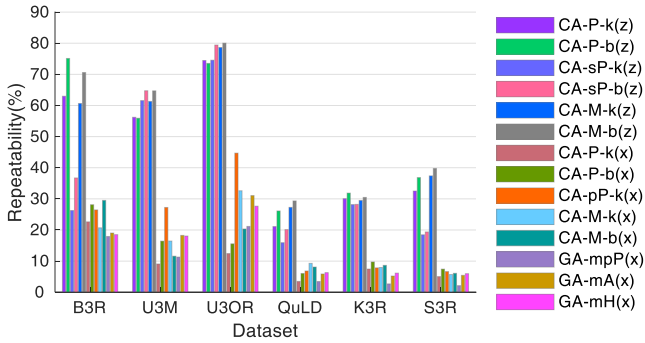

Fig. 7. The repeatability performance of the fourteen axis methods tested on the six experimental datasets.

Among the x-axis methods, several finds can be witnessed. *First*, CA-pP-k(x) achieves the best performance on U3M and U3OR datasets, and outperforms the others by a large margin. This is because the weight of the height along z-axis (i.e., $w_h$) used in covariance analysis can significantly improve the robustness to occlusion as verified in Sect. IV-B. *Second*, for the two mesh-based methods (i.e., CA-M-k(x) and CA-M-b(x)), they both achieve superior performance on the three low-quality datasets (i.e., QuLD, K3R and S3R). In addition, CA-M-b(x) achieves the best performance on B3R dataset, and CA-M-k(x) achieves superior performance on B3R and U3OR. This can verify that the mesh-based x-axes are robust to the scene with comprehensive nuisances. *Third*, for the three GA-based methods, the GA-mA(x) and GA-mH(x) show a superior performance on U3M and U3OR, while exhibit an inferior performance on B3R It is because using a single point for determining x-axis is robust to occlusion, while vulnerable to other nuisances (e.g., noise and varying mesh resolutions). The GA-mpP(x) is the worst x-axis method, exhibiting the worst performance on the three low-quality datasets (i.e., QuLD, K3R and S3R) and B3R dataset.

### F. Robustness

The robustness of all the fourteen axis methods are comprehensively evaluated from eight aspects: the robustness to Gaussian noise, varying mesh resolutions, distance to boundary, keypoint location error, occlusion, clutter and partial overlap. The result of this evaluation is exhibited in Fig. 8. Overall, the performance of the z-axis methods is obviously superior to that of the x-axis methods in terms of the robustness to all the four nuisances. It is because the x-axes are more vulnerable to various nuisances than the z-axes.

*1) Gaussian Noise*

As presented in Fig. 8(a), two main observations can be found. *First*, in the comparison of all the z-axes, CA-P-b(z) has the strongest robustness to noise followed by the CA-M-b(z) and CA-P-k(z), and the CA-sP-b(z) and CA-sP-k(z) are the two most vulnerable methods to noise. It proves that the barycenter has stronger robustness to noise than the keypoint in constructing covariance matrix, and a small scale of local region is very sensitive to noise, which are somewhat not surprising since the stability of a keypoint and a small scale of local region are easily affected by the drifted points, i.e., noise. *Second*, among all the x-axes, CA-P-b(x) and CA-P-k(x) are the two best methods, and GA-mA(x) and GA-mH(x) are the two worst methods, which can verify that a set of local points has stronger robustness to noise than a single point (e.g., salient point). The reason is that Gaussian noise easily causes the error of a single point.

*2) Varying Mesh Resolutions*

As exhibited in Fig. 8(b), several finds can be witnessed. *First*, in terms of the z-axes, CA-M-b(z) and CA-M-k(z) have the strongest robustness to varying mesh resolutions; CA-P-b(z) and CA-P-k(z) presents a mediocre performance; CA-sP-b(z) and CA-sP-k(z) show the worst robustness because a smaller local region is easily affected by this nuisance. The strongest robustness of CA-M-b(z) and CA-M-k(z) is because the local mesh in constructing covariance matrix employs all information of a local region rather than only the vertex information as used in these points-based methods (e.g., CA-P-b(z) and CA-sP-k(z)). *Second*, among the x-axes, CA-M-k(x) and CA-M-b(x) are the two best methods, outperforming the others by a large margin. This further verify that local mesh in covariance matrix construction has strong robustness to varying mesh resolutions. The CA-pP-k(x) achieves a mediocre performance, and the remaining x-axes exhibit an inferior robustness to varying mesh resolutions. The reason of CA-pP-k(x) superior to other x-axes (except CA-M-k(x) and CA-M-b(x)) is that the weight relevant to the height along z-axis (i.e., $w_h$) used in covariance analysis can improve the robustness to occlusion as verified in Sect. IV-B.

*3) Shot Noise*

As showed in Fig. 8(c), several observations can be given. *First*, among all z-axes, CA-sP-b(z), CA-P-b(z), CA-sP-k(z) and CA-sP-k(z) achieve the strongest robustness to shot noise. The CA-M-b(z) and CA-M-k(z) are the two worst z-axes, obviously inferior to the others. It is because the shot noise causes a large error in mesh construction, resulting in the inferior performance for these mesh-based methods (e.g., CA-M-b(z) and CA-M-k(z)). *Second*, in terms of all x-axes, CA-P-

b(x) and CA-P-k(x) achieve the best performance, outperforming other x-axis methods by a large margin. In contrast to the inferior or mediocre performance of GA-mpP(x) for resisting other nuisances, it achieves a superior robustness to shot noise. The reason is that the shot noise generated along normal has little influence to the value in xy-plane. Similar to the mesh-based z-axes, the mesh-based x-axes (i.e., CA-M-b(x) and CA-M-k(x)) also exhibit the worst performance among all x-axis methods.

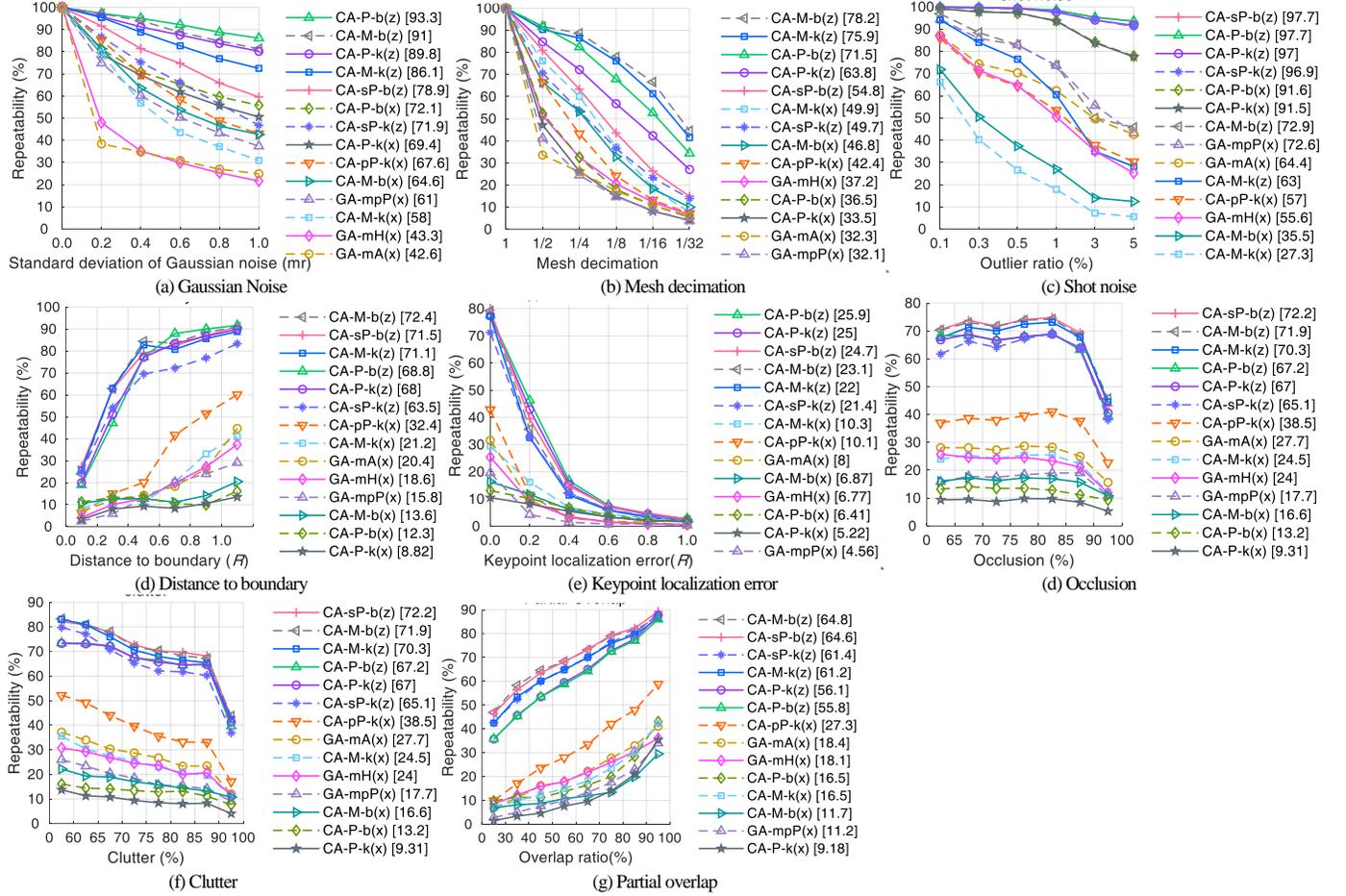

Fig. 8. The repeatability of all the fourteen axis methods with respect to different levels of the eight nuisances.

### 4) Distance to Boundary

Fig. 8(c) presents the result of this test. Among all the z-axes, two observations are given. *First*, CA-M-b(z), CA-sP-b(z) and CA-M-k(z) achieve a superior repeatability on the keypoints close to boundary. This is because a small scale of local region in constructing these axes is robust to occlusion. Note that the support radii for CA-M-b(z) and CA-M-k(z) are both set to 5mr as presented in Table IV. *Second*, CA-sP-k(z) has the worst repeatability on the keypoints away from boundary, obviously inferior to the others. It verifies that the keypoint is obviously inferior to the barypoint in constructing CA-sP-based z-axes.

In the comparison of all x-axes, two finds can be presented. *First*, CA-pP-k(x) achieves the best performance with the distance to boundary larger than 4mr, while its performance obviously drops along with the distance to boundary from 20 to 8mr. This verifies that CA-pP-k(x) is easily affected by missing region. *Second*, CA-M-b(x) and CA-P-b(x) have an inferior repeatability, while present a stable performance with respect to varying distances to boundary, which indicates that the main factors for limiting their performance are not missing region.

### 5) Keypoint Localization Error

Fig. 8(e) exhibits this test result. In terms of all z-axes, the repeatability gaps of them are small, and they all have a significant drop along with the increase of keypoint localization error. CA-P-b(z) achieves the relatively better result, and CA-sP-k(z), CA-M-k(z) and CA-M-b(z) are slightly inferior to the others.

Regarding all x-axes, two views can be given. *First*, CA-M-k(x), CA-pP-k(x), GA-mA(x), GA-mH(x) and GA-mpP(x) present an obvious drop along with the increase of keypoint localization error from 0.0 to 4mr, which indicates that they are sensitive to this nuisance. *Second*, CA-M-k(x) shows a superior performance under no keypoint localization error, and also show a slower performance drop than the above five x-axis methods with the increase of keypoint localization error. This can verify that CA-M-k(x) has strong robustness to keypoint localization error.

### 6) Occlusion

As shown in Fig. 8(e), the repeatability of all fourteen axis methods presents a similar variation trend as the increase of occlusion. Specifically, they exhibit a stable performance as the occlusion increased from 0% to 85%, while present an obvious drop of repeatability as the further increase of occlusion. Among all z-axes, CA-sP-b(z), CA-M-b(z) and CA-M-k(z) present the three best performance, outperforming other z-axis methods under the occlusion larger than 85%. It indicates these three methods has stronger robustness to occlusion than the others. Regarding all x-axes, CA-pP-k(x) has obviously superior performance to other x-axis methods under all levels of occlusion, verifying its stronger robustness to occlusion than other x-axis methods.

*7) Clutter*

As exhibited in Fig. 8(f), the repeatability of all fourteen axis methods presents a slight drop as the clutter increasing from 0% to 90%, while significantly deteriorate under the further increase of clutter, indicating that these axis methods are sensitive to strong clutter. All z-axis methods present a similar performance, and in these z-axis methods, CA-sP-k(z) slightly shows a more drop of repeatability than the others with the increase of clutter. Among all x-axes, similar to the robustness to occlusion, CA-pP-k(x) and CA-P-k(x) consistently achieves the best and the worst repeatability under all levels of clutter, respectively.

*8) Partial Overlap*

The result of this evaluation is shown in Fig. 8(f). Among all z-axes, CA-M-b(z) and CA-sP-b(z) consistently achieve the the two best performance under all levels of overlap ratios, and CA-P-b(z) and CA-P-k(z) are very sensitive to low overlap ratio. It verifies that using a small scale of local region combining with the barycenter as the center in covariance analysis can improve the robustness to low overlap ratios. Note that the support radius for CA-M-b(z) is set to 5mr as shown in Table IV. In terms of all x-axes, although CA-pP-k(x) has obvious higher repeatability than the others, it exhibits a significant drop as the decrease of overlap ratio.

*G. Feature Matching Performance*

A LRF or LRA (i.e., the z-axis of a LRF) is the basis for constructing most local feature descriptors. The feature matching result of three LRA-based (i.e., Spin Image [39], LFSH [40], MaSH [24]) and three LRF-based (SHOT [41], RoPS [6] and TOLDI [9]) descriptors generated on different LRAs or LRFs is presented in Fig. 9. Note that the LRFs are produced by combining the eight x-axis methods with CA-sP-b(z) which has high repeatability on U3M dataset.

Regarding the performance of the LRA-based descriptors (Fig. 9(a)), the performance rank of all z-axis methods in constructing each descriptor is similar. Specifically, the CA-sP-b(z)-based descriptors achieve the best performance followed by the CA-sP-k(z)-based descriptors, and the CA-P-b(z)-based descriptors show the worst performance. This rank is the same with the repeatability rank of these z-axis methods tested on U3M as detailed in Sect. IV-E, which verifies that the repeatability of these z-axis methods is positively correlative to the performance of these LRA-based descriptors. In the comparison of the three LRA-based descriptors, MaSH achieves the best performance. It is because MaSH has higher descriptiveness and stronger robustness than Spin image and LFSH [19].

For the result of the LRF-based descriptors (Fig. 9(b)), two observations can be made. *First*, the differences of different x-axis methods in constructing each LRF-based descriptor are more obvious than the differences of different z-axis methods in generating each LRA-based descriptor, which indicates that these LRF-based descriptors are sensitive to the repeatability of x-axis. *Second*, the performance rank of these x-axis methods in constructing each descriptor is similar. Specifically, CA-pP-k(x)-, GA-mH(x)- and GA-mA(x)-based descriptors achieve the best performance; CA-M-k(x)- and S_pP(x)-based ones present a moderate performance; CA-P-k(x)-, CA-P-b(x)- and CA-M-b(x)-based ones are the three worst methods. This rank is similar with the repeatability rank of these x-axis methods tested on U3M as detailed in Sect. IV-E, verifying that the performance of these LRF-based descriptors is positively correlative to the repeatability of these x-axis methods.

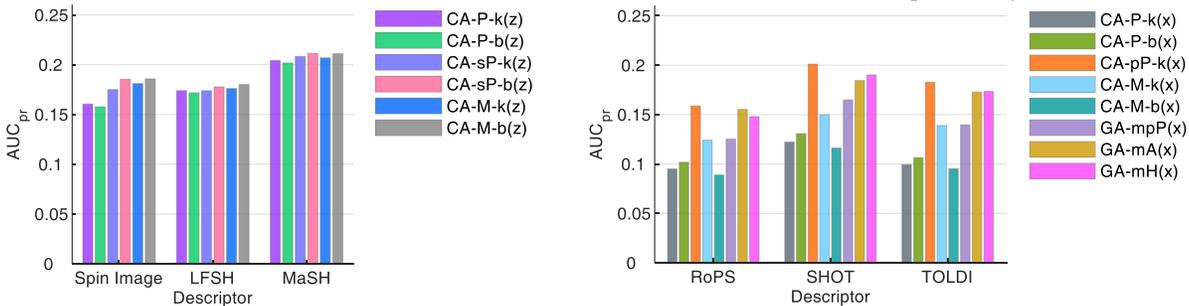

(a) The performance of the LRA-based descriptors with respect to different z axis methods

(b) The performance of the LRF-based descriptors with respect to different LRFs constructed by different x-axis methods combining with CA-sP-b(z)

Fig. 9. The AUC$_{pr}$ performance of three LRA-based and three LRF-based descriptors with respect to different axis methods tested on U3M dataset.

*H. Transformation Estimation Performance*

LRF and LRA are the important tools for designing transformation estimation methods [15, 22]. The result of different z-axis methods in constructing two LRA-based transformation estimation methods (i.e., 2P-RANSAC[24] and CG-SAC[42]) and different x-axis methods in generating three LRF-based ones (i.e., CCV[44], 1P-RANSAC[43] and LRF-MCS[26]) is shown in Fig. 10.

For the result of LRA-based transformation estimation methods (Fig. 10(a)), the performance gaps of all z-axis methods in designing 2P-RANSAC and CG-SAC are small. It is because the main factor for restricting their performance is the correctness of point-to-point correspondences. Regarding the result of the LRF-based methods as exhibited in Fig. 10(b), the performance rank of all x-axis methods in designing each transformation estimation method is similar. Specifically, CA-pP-k(x)-, GA-mH(x)- and GA-mA(x)-based descriptors achieve the best performance; CA-M-k(x)- and GA-mpP(x)- based ones present a moderate performance; CA-P-k(x)-, CA-P-b(x)- and CA-M-b(x)-based ones are the three worst methods. It verifies that the performance these LRF-based transformation estimation methods is positively correlative to the repeatability of these x-axex. Contrast to the stable performance of the LRA-based transformation estimation method with respect to varying z-axis methods, the performance fluctuation of each LRF-based one is more obvious combined with different x-axes.

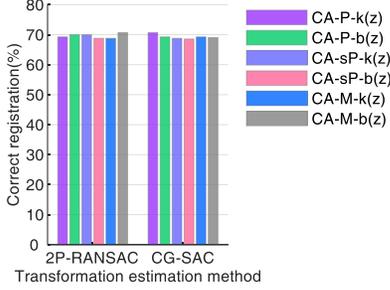
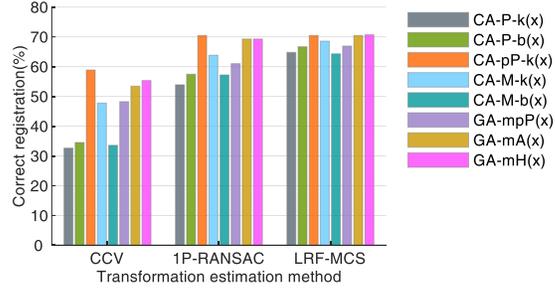

(a) The performance of the two LRA-based transformation estimation methods with respect to different z-axis methods

(b) The performance of the three LRF-based transformation estimation methods with respect to different LRFs generated by different x-axis methods with CA-sP-b(z)

Fig. 10. The percentages of correct registration for two LRA-based and three LRF-based transformation estimation methods with respect to different axis methods tested on U3M dataset.

*I. Computational Efficiency*

Fig. 11 exhibits the time costs of all the fourteen methods for constructing an LRF axis with respect to varying support radii. Several observations can be made. *First*, the four mesh-based methods (i.e., CA-M-k(z), CA-M-b(z), CA-M-k(x) and CA-M-b(x)) are very time-consuming. Their efficiencies are inferior to the others by a large margin. This is because the covariance analysis of these mesh-based methods is implemented by using an entire local surface rather than only the vertices of the surface. *Second*, the three GA-based methods (i.e., GA-mH, GA-mpP and GA-mA) are very efficient. The reason is that these methods only need to search a salient point or average the projected local points, without conducting covariance analysis. *Third*, the keypoint-based method is generally more efficient than the barycenter-based counterpart. It is because the keypoint is directly available, while the barycenter needs to calculate the average of local points. *Fourth*, among all local points-based methods, the CA-pP-k(x) is slightly inferior to the others in terms of efficiency. This is because it needs to project local points on xy-plane before implementing covariance analysis.

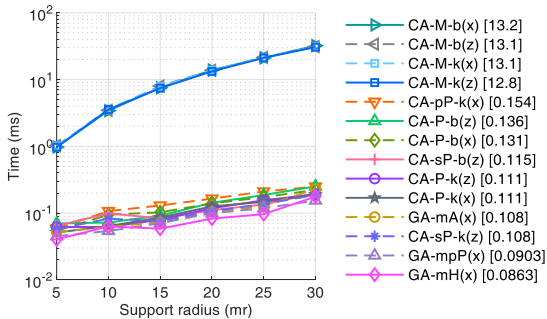

Fig. 11. The time costs of all evaluated axis methods for generating an LRF axis with respect to varying support radii.

## V. SUMMARY AND DISCUSSION

Based on the tested results, the superior and inferior x- and z-axis methods with respect to different test items are reported in Table IX. Several findings are summarized as follows.

i) In terms of the performance with different support radii, a large support radius is robust to noise and varying mesh resolutions as tested on B3R, while vulnerable to occlusion and clutter as tested on U3M and U3OR; the x-axis methods are commonly more stable than the z-axis ones for resisting occlusion along with the increase of support radius.

ii) The optimal weight for these LRF axes are different. Specifically, assigning each neighbor point the same weight value (i.e., $w_0$) is more appropriate for the barycenter-based methods (e.g., CA-P-b(z) and CA-M-b(z)), and $w_r$ is more appropriate for the keypoint-based ones (e.g., CA-P-k(z) and CA-M-k(z)); the $w_a$ is an effective weight for improving the performance of mesh-based methods, and the $w_h$ is a very effective weight strategy for generating x-axis (e.g., CA-pP-k(x) and GA-mpP(x)).

iii) In terms of disambiguation, the normal mean-based method is commonly superior to the points mean-based one. Therefore, the normal mean-based method is recommended to disambiguate the CA-based LRF axes.

iv) In the comparison between the z-axes and the x-axes, the repeatability of x-axes are obviously inferior to that of the z-axes.

v) For all z-axis methods, the CA-M-b(z) has the best overall repeatability and robustness on all the test items; the covariance matrix constructed by using a small scale of local region (e.g., CA-sP-b(z)) is robust to occlusion, while very vulnerable to noise and varying mesh resolutions; the barycenter-based z-axis methods (e.g., CA-P-k(z), CA-sP-k(z) and CA-M-k(z)) are commonly superior to the keypoint-based countparts; the mesh-based methods (e.g., CA-M-b(z)) are very robust to varying mesh resolutions, while vulnerable to shot noise and very time-consuming.

vi) Regarding all x-axis methods, CA-pP-k(x) has the strongest robustness to occlusion and clutter, and therefore be suitably applied in the scene with the main nuisance being occlusion and clutter (e.g., U3M and U3OR datasets).

Choosing a single point to determine (e.g., GA-mA(x) and GA-mH(x)) x-axis is robust to occlusion and clutter, while susceptible to Gaussian noise and keypoint localization error. Local point-based x-axis methods have strong robustness to noise, and the mesh-based ones are very robust to varying mesh resolutions.

TABLE IX
SUPERIOR AND INFERIOR AXIS METHODS IN DIFFERENT ASPECTS OF THE EXPERIMENTAL COMPARISON

| | | z-axis method | | x-axis method | |
|---|---|---|---|---|---|
| | | Superior | Inferior | Superior | Inferior |
| Repeatability | B3R | CA-P-b(z), CA-M-b(z) | CA-sP-k(z), CA-sP-B(z) | CA-M-b(x), CA-P-b(x), CA-pP-k(x) | GA-mpP(x), GA-mA(x), GA-mH(x) |
| | U3M | CA-sP-b(z), CA-M-b(z) | CA-P-k(z), CA-P-b(z) | CA-pP-k(x) | CA-P-k(x), CA-M-b(x), GA-mpP(x) |
| | U3OR | CA-sP-b(z), CA-M-b(z), CA-M-k(z) | CA-P-k(z), CA-P-b(z), CA-sP-k(z) | CA-pP-k(x) | CA-P-k(x), CA-P-b(x) |
| | QuLD | CA-M-b(z), CA-M-b(z), CA-P-b(z) | CA-sP-k(z) | CA-M-k(x), CA-M-b(s) | CA-P-k(x), GA-mpP(x) |
| | K3R | CA-P-b(z) | CA-sP-k(z), CA-sP-b(z) | CA-P-b(x) | GA-mpP(x) |
| | S3R | CA-M-b(z) | CA-sP-k(z), CA-sP-b(z) | CA-P-b(x), CA-pP-k(x) | GA-mpP(x) |
| Robustness | Gaussian noise | CA-P-b(z) | CA-sP-b(z), CA-sP-k(z) | CA-P-b(x), CA-P-k(x) | GA-mH(x), GA-mA(x) |
| | Mesh decimation | CA-M-b(z), CA-M-k(z) | CA-sP-k(z), CA-sP-b(z) | CA-M-k(x), CA-M-b(x) | GA-mpP(x), GA-mA(x), CA-P-k(x) |
| | Shot noise | CA-sP-b(z), CA-P-b(z), CA-P-k(z) | CA-M-k(z), CA-M-b(z) | CA-P-b(x), CA-P-k(x) | CA-M-k(x), CA-M-b(x) |
| | Distance to boundary | CA-M-b(z), CA-sP-b(z), CA-M-k(z) | CA-P-K(z), CA-P-b(z) | CA-pP-k(x) | CA-P-k(x), GA-mH(x) |
| | Keypoint localization error | CA-P-b(z) | CA-sP-k(z), CA-M-k(z) | CA-M-k(x) | GA-mpP(x), GA-mH(x), GA-mA(x), CA-pP-k(x) |
| | Occlusion | CA-sP-b(z), CA-M-b(z) | CA-sP-k(z) | CA-pP-k(x) | CA-P-k(x) |
| | Clutter | CA-sP-b(z), CA-M-b(z) | CA-sP-k(z) | CA-pP-k(x) | CA-P-k(x) |
| | Partial overlap | CA-M-b(z), CA-sP-b(z) | CA-P-b(z), CA-P-k(z) | CA-pP-k(x) | CA-P-k(x), GA-mpP(x) |
| Feature matching | | CA-sP-b(z), CA-M-b(z) | CA-P-b(z), CA-P-k(z) | CA-pP-k(x), GA-mA(x), GA-mH(x) | CA-M-b(x), CA-P-k(x), CA-P-b(x) |
| Transformation estimation | | CA-P-k(z), CA-sP-k(z) | CA-sP-b(z) | CA-pP-k(x), GA-mA(x), GA-mH(x) | CA-M-b(x), CA-P-k(x), CA-P-b(x) |
| Efficiency | | CA-sP-k(z), CA-P-k(z) | CA-M-b(z), CA-M-k(z) | GA-mH(x), GA-mpP(x), GA-mA(x) | CA-M-b(x), CA-M-k(x) |

## VI. CONCLUSION

In this paper, a comprehensive evaluation of the z-axis, x-axis, weight and disambiguation methods in LRF construction has been conducted on six benchmark datasets with different nuisances (e.g., noise, occlusion, clutter and varying mesh resolutions) and scanners (including Minolta vivid, Kinect and Space Time). Specifically, this paper firstly evaluated and set the appropriate parameters for all tested methods, including support radius, weight, disambiguation and the appropriate z-axes for constructing these z-axis-dependent x-axes. Based on these appropriate parameters, the repeatability of all LRF axes are comprehensively evaluated on six datasets. The robustness of these LRF axes to eight nuisances (including Gaussian noise, varying mesh resolutions, shot noise, distance to boundary, keypoint localization error, occlusion and clutter) were comprehensively evaluated. In addition, this paper also evaluated the performance of these LRF axes in the feature matching and transformation estimation. Finally, the merits and demerits of these methods were summarized and discussed. To the best of our knowledge, this is the first work for comprehensively evaluating the z- and x-axis methods as well as the weight and disambiguation strategies in LRF construction. Contrast to the evaluation of complete LRF methods, this evaluation strategy can reflect the performance of LRF more accurately and comprehensively. This paper thus can be regarded as a guide for users to select the most appropriate LRF or LRA and for developers to design a more superior LRF axis.

## ACKNOWLEDGMENT

The authors would like to acknowledge the Stanford 3D Scanning Repository, the University of Western Australia and the University of Bologna for making their datasets available to us.